\documentclass[letterpaper, 10 pt, conference]{ieeeconf}
\usepackage{graphicx} %

\IEEEoverridecommandlockouts                              %
\usepackage{times} %
\usepackage{amsmath} %
\usepackage{amssymb}  %
\usepackage[hidelinks]{hyperref}
\usepackage{comment}
\usepackage{tabularx}
\usepackage{subcaption}
\usepackage{hhline}
\usepackage{xspace}
\usepackage{tikz}
\usepackage{float}
\usepackage{pgfplots}
\usepackage{booktabs}
\usepackage{balance}
\newcommand{\labeltarget}[1]{\Hy@raisedlink{\hypertarget{#1}{}}}

\usepackage{cite}

\usepackage[font=small,labelfont=bf]{caption}
\usepackage{color}
\usepackage{xcolor}

\newcommand{\mcN}{\mathcal{N}}
\newcommand{\mcL}{\mathcal{L}}

\newcommand{\subscr}[2]{#1_{\text{#2}}}

\newcommand\mydots{\hbox to 1em{.\hss.\hss.}}
\newcommand{\until}[1]{\{1, \mydots,#1\}}
\newcommand{\untilwithzero}[1]{\{0, \mydots,#1\}}

\newcommand{\ocp}{\hyperref[eq:ocp]{\textbf{OCP}}\xspace}
\newcommand{\infoocp}{\hyperref[eq:info-ocp]{\textbf{Info-OCP}}\xspace}

\newcommand{\modelone}{\text{Once-OCP-Model}\xspace}
\newcommand{\modeltwo}{\text{Once-Info-Model}\xspace}
\newcommand{\modelthree}{\text{Twice-Info-Model}\xspace}

\newcommand{\hwplotRed}{\raisebox{2pt}{\tikz{\draw[red,solid,line width=1pt](0,0) -- (4mm,0);}}}
\newcommand{\hwplotBlue}{\raisebox{2pt}{\tikz{\draw[blue, solid,line width=1pt](0,0) -- (4mm,0);}}}

\newcommand{\real}{\mathbb{R}}

\let\labelindent\relax
\usepackage{enumitem} %

\title{\LARGE \bf
First, Learn What You Don’t Know:\\Active Information Gathering for Driving at the Limits of Handling
}

\author{Alexander Davydov$^{1,2}$, Franck Djeumou$^{2,3}$, Marcus Greiff$^{2}$, \\
Makoto Suminaka$^{2}$,  Michael Thompson$^{2}$, John Subosits$^{2}$,  and Thomas Lew$^{2}$
\thanks{$^*$ The work of A. Davydov was carried out at Toyota Research Institute.}
\thanks{$^{1}$ %
Department of Mechanical Engineering, Rice University, Houston, TX, USA}
\thanks{$^{2}$ %
Toyota Research Institute, Los Altos, CA, USA}
\thanks{$^{3}$ %
MANE, Rensselaer Polytechnic Institute, Troy, NY, USA}
}

\begin{document}
\thispagestyle{empty}
\pagestyle{empty}

\maketitle

\begin{abstract}
Combining %
data-driven models that adapt online and model predictive control (MPC) has enabled effective control of nonlinear systems. %
However, when deployed on unstable systems, online adaptation may not be fast enough to ensure reliable simultaneous learning and control. 
For example, a controller on a vehicle executing highly dynamic maneuvers—such as drifting to avoid an obstacle—may push the vehicle's tires to their friction limits, destabilizing the
vehicle and allowing modeling errors to quickly compound and
cause a loss of control.
To address this challenge, 
we present an active information gathering framework for identifying vehicle dynamics as quickly as possible.
We propose an expressive vehicle dynamics model that leverages Bayesian last-layer meta-learning to enable rapid online adaptation. 
The model's uncertainty estimates are used to guide informative data collection and quickly improve the model prior to deployment. 
Dynamic drifting experiments on  a Toyota Supra show that (i) the framework enables reliable control %
of a vehicle at the edge of stability, (ii) online adaptation alone may not suffice for zero-shot control and can lead to undesirable transient errors or spin-outs, and 
(iii) 
active data collection helps achieve reliable performance. 
\end{abstract}

\section{Introduction}
Controlling unstable nonlinear systems remains challenging \cite{Stein2003}. For example, %
driving a vehicle through dynamic and unstable drifting maneuvers 
is a difficult task where 
slight decision-making mistakes 
can lead to spin outs or crashes. In recent years, 
advanced control
methods 
have been developed 
to give autonomous vehicles the 
ability to leverage their full handling potential 
\cite{velenis2011steady,cutler2016autonomous,djeumou2023autonomous,goh2024beyond,Weber24}, with possible applications to the future design of advanced driver-assistance systems and autonomous driving safety systems. 

In this setting, data-driven model predictive control (MPC) methods    
have gained wide popularity \cite{Hewing2020} due to their high performance, capability to use expressive learned models, and ability to account for constraints such as actuator limits and obstacle avoidance. 
In particular, data-driven models that adapt using online data can reduce model errors and sim-to-real discrepancies 
at deployment time \cite{kabzan2019learning,spielberg2021neural,djeumou2024one,ding2024drifting}.  
However, simultaneously controlling and learning unstable systems is notoriously difficult. 
For instance (see Sec.~\ref{sec:results}), successfully executing a drifting maneuver requires a sufficiently accurate initial model %
to precisely initiate the drift, and online adaptation alone may not 
be sufficient to execute the maneuver. %

Autonomous driving at the limits presents a challenge for data-driven MPC with online adaptation due to its fast-paced nature and the inherent instability of the executed maneuvers.
These difficulties raise the following questions:
How can we effectively identify uncertain nonlinear dynamics to ensure that subsequent control is reliable? 
How should informative data be collected efficiently and safely to minimize time-intensive and potentially expensive testing on hardware?

\begin{figure}
    \includegraphics[width=1\linewidth]{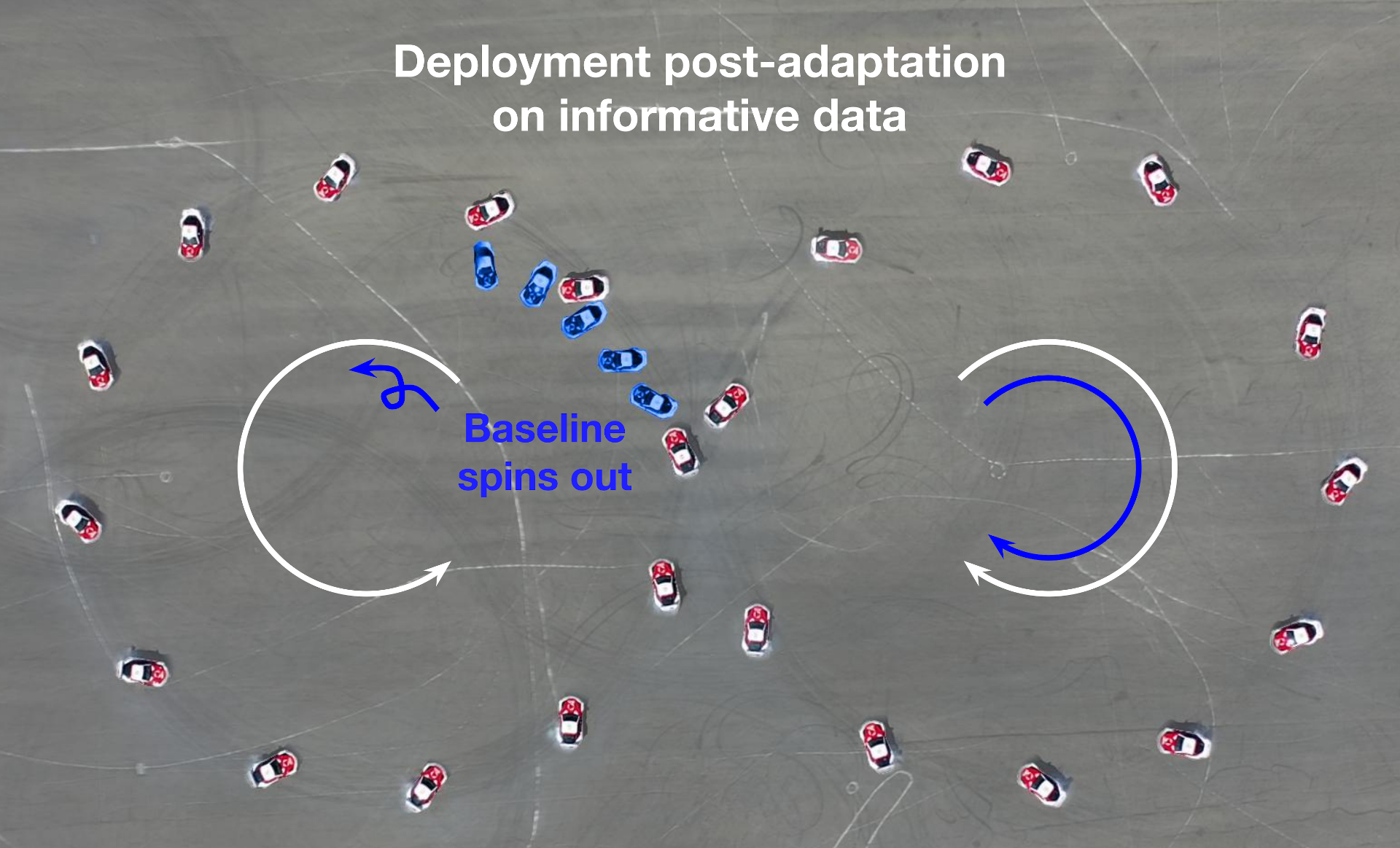}
    \caption{We propose an active information gathering framework that guides data collection to quickly learn a model in a controlled (i.e., safe) setting. Adapting on this information-rich data enables reliable control in dynamic drifting trajectories. In contrast, directly using the model without adapting prior to deployment may lead to unreliable control performance, such as spinning out.
    }\label{fig:main:figure8}
\end{figure}

\textbf{Contributions}: We present an active information gathering framework to quickly identify nonlinear dynamics for reliable subsequent deployment in MPC. Specifically:
\begin{itemize}[leftmargin=3.5mm]
\item We present a data-driven dynamics model tailored for driving at the limits of handling.
This model combines prior physics knowledge with learned expressive neural network features.
By adapting only the last layer, this data-driven model enables rapid online adaptation, significantly improving upon non-adaptive models in controlling unstable systems where small initial errors can quickly amplify.
\item To enable reliable deployment of the learned model in an MPC framework,  we propose a pre-deployment active learning approach to identify an accurate model as fast as possible while satisfying constraints during data collection. 
\item We extensively validate the proposed approach on a Toyota Supra in two different dynamic drifting maneuvers. 
In particular, results show that online adaptation of the meta-trained model alone may be insufficient to initiate a drift reliably and can lead to large transients in tracking error.
On the other hand, the proposed active information gathering approach enables rapid identification of the system's dynamics and yields improved tracking performance.
\end{itemize} 
From a vehicle control perspective, our results demonstrate that 
executing drifting maneuvers 
is possible by adapting a model on %
informative non-drifting trajectories. 
Specifically, information-rich trajectories do not necessarily involve drifting, %
and adapting on such data significantly improves the execution of challenging drifting maneuvers.

\section{Related Work}

Data-driven models are increasingly popular 
for control due to their ability to leverage expressive models and adapt them online given measurements from the system, see \cite{Hewing2020} and \cite{Brunke2022} for recent reviews. For example, model predictive control (MPC) approaches leveraging neural network models have been widely used in recent years for dynamic driving at the limits of handling  \cite{spielberg2021neural,djeumou2024one,ding2024drifting,djeumou2023autonomous}. These approaches adapt models online %
with the implicit assumption that the model will adapt fast enough during the control task to yield satisfactory performance, which may be difficult for systems where small initial errors quickly amplify. %

\textit{Online learning} and control algorithms that \textit{actively} seek informative data to improve the model \cite{Abbasi2011,wagenmaker2024optimal,Koller2018,Buisson2020,mania2022active,saviolo2023active,Zhang2021,lew2022safe} often result in improved control performance compared to passively adapting a model while performing a task, and allow enforcing safety constraints during the data collection process \cite{Koller2018,lew2022safe}. %
Leveraging such active learning approaches for controlling unstable nonlinear systems remains challenging. Indeed, the tradeoffs between seeking informative data (exploration) and performing the task (exploitation) are well-known;
unstable systems  
are particularly difficult to simultaneously learn and control \cite{Dean2019,Mania2019}. 
These tradeoffs motivate the design of a framework where active learning is done \textit{prior to deployment} in a controlled setting to learn a model for control with minimal experimentation. 

Autonomous driving at the limits of handling exemplifies these tradeoffs. In this application,  the vehicle may 
operate with saturated tire forces close to the road boundaries, and 
it is desirable to identify the vehicle dynamics as accurately as possible 
prior to deployment. 
Previous approaches to driving at the limits  leverage  MPC with 
physics-based~\cite{goh2024beyond,Weber24} and neural network-based~\cite{spielberg2021neural,djeumou2023autonomous,ding2024drifting} dynamics models. Recently, \cite{djeumou2024one} introduced a neural SDE dynamics model for MPC where the parameters of the SDE are sampled from a diffusion model conditioned on online observations, enabling \textit{passive} online adaptation during task execution. 
Our proposed active learning framework complements
these works, enabling the identification of uncertain dynamics as fast
as possible prior to deployment in model-based control.

Active learning approaches to nonlinear system identification include methods that maximize information gain \cite{Buisson2020,lew2022safe,kim2023bridging,sukhija2024optimistic}. 
Our approach relies on similar information-theoretic ideas to identify uncertain dynamics as quickly as possible. 
We leverage a Bayesian modeling approach \cite{harrison2020meta} to last-layer meta-learning \cite{lee2019meta,bertinetto2019meta,harrison2020meta,Richards2023,o2022neural} that is conducive to information gathering \cite{lew2022safe}. Compared to driving at the limits of handling, experiments in \cite{Buisson2020,lew2022safe,sukhija2024optimistic} involve simulated or slow-paced robotic systems 
that do not face the same challenges of
controlling and learning unstable dynamics. 
\cite{kim2023bridging} also study active exploration for off-road driving, but use probabilistic ensemble neural networks that are computationally expensive compared to last-layer meta-trained neural network models, and penalize the vehicle sideslip which prevents executing drifting maneuvers.
Additionally, as we show in Sec.~\ref{sec:results}, online adaptation alone may not suffice  to reliably execute drifting maneuvers, even using a meta-learned model capable of rapid adaptation.

%
%
%

%

%

%
%
%
\section{Bayesian vehicle dynamics model}\label{sec:model}
\begin{figure}
    \centering
    \includegraphics[width=0.775\linewidth]{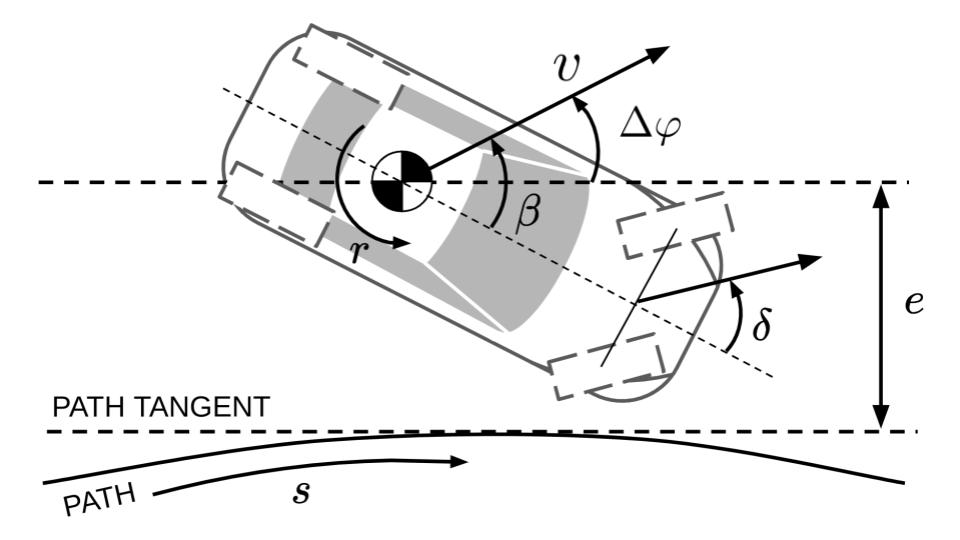}
    \vspace{-7pt}
    
    \caption{Curvilinear coordinate system of the vehicle, from \cite{ding2024drifting}.
    }
    \label{fig:coordinates}
\end{figure}

Central to our approach is a vehicle dynamics model that is expressive, captures uncertainty, and enables rapid online adaptation. To this end, we combine a physics-based nominal bike model with a Bayesian last-layer meta-learning model. The former encodes prior knowledge about the vehicle, while the latter captures model mismatch and enables rapid adaptation over expressive neural network features.  The vehicle state is expressed in curvilinear coordinates \cite{goh2024beyond,ding2024drifting}, see Fig.~\ref{fig:coordinates}. As in \cite{goh2024beyond,ding2024drifting}, we do not use brake inputs in this work. The states and control inputs of the vehicle are
\begin{align}\label{eq:statedef}
x&=
(r, 
v,
\beta,
\omega_r,
e,
\Delta\varphi)\in\mathbb{R}^n,
\quad 
u=(\delta,\tau)\in\mathbb{R}^m,
\end{align}
where $(n,m)=(6,2)$ are the state and control dimensions, $r$ is the yaw rate, $v$ is the total planar velocity, $\beta$ is the sideslip, $\omega_r$ is the rear wheelspeed, $e$ and $\Delta\varphi$ are the lateral and angle deviations to the reference trajectory, $\delta$ is the steering angle, and $\tau$ is the engine torque, respectively.

The vehicle is modeled with discrete-time dynamics as %
\begin{equation}\label{eq:dynamics}
x_{k+1}=f_\xi(x_k,\tilde{u}_k)+\epsilon_k,
\end{equation}
where $k\in\mathbb{N}$ and %
$\epsilon_k\sim\mcN(0, \text{diag}(\sigma_1^2, \dots, \sigma_n^2))$ are Gaussian-distributed disturbances that are independent over time $k$ and state dimensions. $\xi$ represents unknown parameters of the dynamics (e.g., corresponding to unmodeled phenomena) and $\tilde{u}_k = (u_k,\dot{u}_k)$ includes the control input rate $\dot{u}_k=(\dot{\delta}_k,\dot{\tau}_k)$ at time $k$. Including $\dot{u}_k$ in $\tilde{u}_k$ and \eqref{eq:dynamics} accounts for the fact that MPC plans $(u_0,u_1,\dots,u_N)$ are executed by a low-level controller via linear interpolation on the control horizon. That is, %
$u(k\Delta t + \tau)=u_k+\tau\dot{u}_k$ with a time step $\Delta t$ and $\dot{u}_k=(u_{k+1}-u_k)/\Delta t$ for $\tau\in[0,\Delta t)$. 

\newcommand\timeind{k}

We approximate the dynamics \eqref{eq:dynamics} using the model
\begin{equation}\label{eq:model}
x_{k+1} = h(x_k,u_k) + g(z_k) + \epsilon_k,
\end{equation}
where $h$ is the $\Delta t$-time step Euler integration of a standard bike model (see Appendix~\ref{app:bike} for details).
The residual $g$ in~\eqref{eq:model} is modeled by
$g(z)=(g_1,\dots,g_n)(z)$ with inputs $$z_k=((r_k,v_k,\beta_k,\omega_{r,k}),u_k,u_{k+1}),$$ and each dimension $i$ of $g(\cdot)$ is linear in the learned features $\phi_i(z)\in\real^d$ for some parameters $\theta_i\in\real^d$, that is
\begin{equation}
    g_i(z_k) := \theta_i^\top \phi_i(z_k), \quad i \in \until{n}.
\end{equation}
We omit the variables $e$ and $\Delta\varphi$ since they do not affect the dynamics of the other state variables,
see~\cite{goh2024beyond}. %
That is, given the state in~\eqref{eq:statedef}, we let $(g_5,g_6)(z)=0$ for all $z$.

The features $\phi_i(\cdot)$ are neural networks that are learned jointly with the parameters $\theta_i$ via \textit{last-layer Bayesian meta-learning} \cite{harrison2020meta,Banerjee2020,lew2022safe} to yield an expressive model capable of rapid adaptation with uncertainty estimates. Specifically, viewing $g_i(z_k)$ as a neural network, the last-layer parameter $\theta_i$ are parameterized using a Gaussian distribution
\begin{equation}\label{eq:theta:gaussian}
\theta_i \sim \mcN(\bar{\theta}_i, \sigma_i^2 \Lambda_i^{-1}), \quad i \in \until{n},
\end{equation}
with mean parameters $\bar{\theta}_i$ and positive definite precision matrices $\Lambda_i/\sigma_i^2$. Following~\cite{lew2022safe}, we model independent Gaussian distributions of each $\theta_i$, such that the joint distribution for the last layer parameters $(\theta_1, \dots, \theta_n)$ are Gaussian distributed. 

\textit{Computing the posterior:} The linear structure of $g_i(z)$ enables the computation of the last-layer parameters $\theta_i$ via Bayesian linear regression. Specifically, given prior distributions $\mcN(\bar{\theta}_{i,0}, \sigma_i^2 \Lambda_{i,0}^{-1})$ for $\theta_i$, and a state-control trajectory $\mathcal{D}_T=\{(z_k,x_{k+1})\}_{k=0}^T$, the posterior $\theta_{i,T}|\mathcal{D}_T$ is Gaussian and can be computed recursively from $Q_{i,0} = \Lambda_{i,0} \bar{\theta}_{i,0}$ by
\begin{subequations}\label{eq:update}
\begin{align}
\Lambda_{i,\timeind+1}^{-1} &= \Lambda_{i, \timeind}^{-1} - \frac{(\Lambda_{i,\timeind}^{-1}\phi_i(z_\timeind))(\Lambda_{i,\timeind}^{-1}\phi_i(z_\timeind))^\top}{1 + \phi_i(z_\timeind)^\top \Lambda_{i,\timeind}^{-1}\phi_i(z_\timeind)}, \label{eq:update-precision}
\\
Q_{i,\timeind+1} &= Q_{i,\timeind} +  (x_{i,\timeind+1} - h_i(x_\timeind,u_\timeind))\phi_i(z_\timeind),
\label{eq:update-Q}
\\
\bar\theta_{i,\timeind+1}&=\Lambda_{i,\timeind+1}^{-1}Q_{i,\timeind+1}.%
\end{align}
\end{subequations}
Moreover, the one-step predictions are Gaussian-distributed, that is, $x_{i,\timeind+1}\,|\,
x_{0:\timeind},\tilde{u}_{0:\timeind}
\sim
\mcN\left(
\mu_{i,{\timeind+1}},\Sigma_{i,{\timeind+1}}
\right)$ with
\begin{subequations}
\label{eq:posterior_predictions}
\begin{align}
\mu_{i,\timeind+1}&=h_i(x_\timeind,u_\timeind)+ \bar\theta_{i,k}^\top\phi_i(z_\timeind),
\\
\Sigma_{i,\timeind+1}&=\sigma_i^2(1 + \phi_i(z_\timeind)^\top \Lambda_{i,k}^{-1} \phi_i(z_\timeind)).
\end{align}
\end{subequations}
With the proposed model, the mismatch between the true dynamics $f$ and the nominal model $h$ through $g$ is learned such that the one-step-ahead predictions are Gaussian by design, facilitating both meta-learning and information gathering.

\textit{Offline Bayesian meta-training:} The prior parameters $(\bar\theta_{i,0},\Lambda_{i,0}^{-1})$, features $\phi_i$, and noise covariance scalings, $\sigma_i^2$, are pre-trained offline to obtain an expressive model that can be adapted online using the update rule \eqref{eq:update}. Specifically, given a dataset of $J$ trajectories corresponding to different parameters $\xi$ of the true system \eqref{eq:dynamics} and a meta-training horizon $T$, we minimize the negative posterior log-likelihood objective
\begin{equation*}
\subscr{\mcL}{nll}(\bar{\theta},\Lambda, \phi,\sigma) = \sum_{j=1}^J \sum_{\timeind=1}^{T} \|x_{\timeind}^j - \mu_\timeind^j\|_{(\Sigma_{\timeind}^j)^{-1}}^2 + \log(\det(\Sigma_{\timeind}^j)),
\end{equation*}
where $\|y\|_A^2 = y^\top A y$ and $(\mu_{\timeind}^j,\Sigma_{\timeind}^j)$ denotes the one-step-ahead prediction computed by~\eqref{eq:update} at a time $k{-}1$ in dataset $j$. The fact that the posterior parameters are obtained via the update rule is key to learning expressive features $\phi$ that favor rapid online adaptation of the model. For further details on Bayesian meta-learning, we refer to  \cite{harrison2020meta,Banerjee2020,lew2022safe}.

\textit{Implementation details}: We process vehicle state data at $10$Hz and thus define the discrete-time dynamics \eqref{eq:dynamics} and the model \eqref{eq:model} with $\Delta t=0.1\mathrm{s}$. We found that the bike model gives poor estimates of wheelspeed dynamics, and thus learn the full wheelspeed dynamics by setting $h_4(x_k,u_k)=0$. We train the model using code from~\cite{Banerjee2020}. 

As a dataset, 
we train on a mix of successful and unsuccessful manual and autonomous drifting data
totaling approximately 19 minutes of driving. 
This data corresponds to $35$ different trajectories, i.e., realizations of $f_\xi$, and contains a total of $11382$ data points. Following~\cite{ding2024drifting}, we additionally augment our dataset with a mirrored set of data by negating the signs of the yaw rate, sideslip, and steering angle to simulate turning in the opposite direction. This doubles our dataset size.

For our architecture, we use a $\tanh$-feedforward neural network for each $\phi_i$ with two hidden layers of width $128$, and the dimension of each $\theta_i$ is $16$. Following~\cite{lew2022safe}, we share weights and biases across each $\phi_i$ except for their last layer weights and biases $(\subscr{W}{out}^i, \subscr{b}{out}^i) \in \real^{16 \times 128} \times \real^{16}$. In total, there are $27012$ trainable parameters for the neural network, prior parameters, $(\bar{\theta}_{i,0},\Lambda_{i,0}^{-1})$, and noise covariance scalings, $\sigma_i^2$. We use a meta-training horizon $T = 10$ and train our model using Adam with an initial learning rate of $10^{-3}$ and an exponential decay rate of $0.9975$ over $1000$ epochs.

\section{Information Gathering MPC Framework}
\begin{figure}
    \centering
    \includegraphics[width=\linewidth]{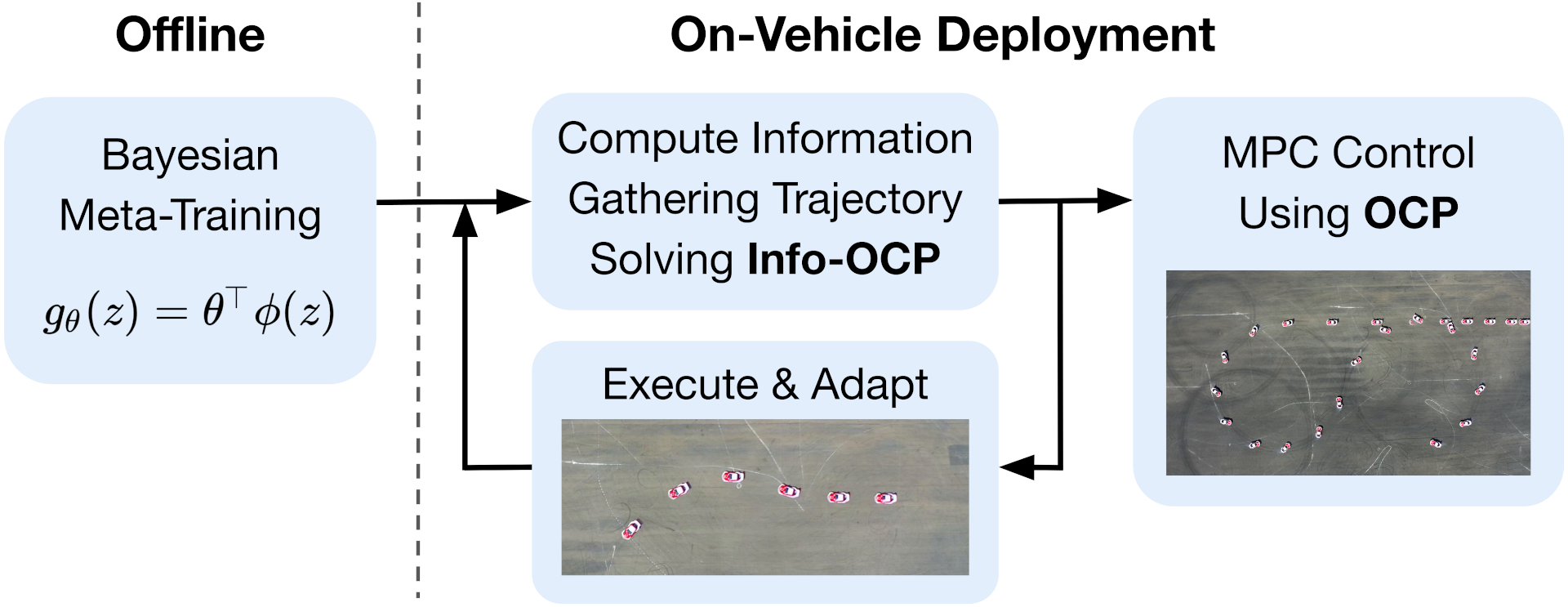}
\vspace{-15pt}

\caption{Proposed active information gathering framework.
\vspace{-3mm}
}
\label{fig:framework}
\end{figure}

We present an active information gathering framework, shown in Fig.~\ref{fig:framework}, to identify nonlinear dynamics as fast as possible for reliable subsequent deployment within MPC. First, we meta-train a Bayesian vehicle model offline (Sec.~\ref{sec:model}) using a dataset of vehicle trajectories. Second, we leverage the model's expressive feature space $\phi$ and uncertainty representation encoded by $(\bar\theta,\Lambda^{-1})$ to safely collect informative data and refine the model. After this information-guided adaptation, we leverage the refined vehicle model within an MPC framework to control the vehicle in challenging driving scenarios such as drifting. Next, we formulate the optimal control problems used for MPC and information gathering.

We define an MPC formulation that enables tracking dynamic trajectories such as drifting maneuvers parameterized by a reference trajectory $x_{\mathrm{ref}}$.  
To favor smooth control inputs, we penalize fast changes in the control inputs and define
\begin{equation*}
\ell_k(x,u,w)
=\|x - x_{\mathrm{ref},k}\|_{Q}^2 + \|u - w\|_{R}^2,
\end{equation*}
where $(Q,R)$ are positive semidefinite diagonal matrices. 
We impose constraints to ensure that the vehicle remains on the track and that actuators operate within hardware constraints. To plan over a horizon $N$ from an initial state $x_{\text{init}}$, we formulate an optimal control problem (\textbf{OCP})
\begin{subequations}\label{eq:ocp}
\begin{align}
\min_{x,u}
\ 
&\textstyle\sum_{k=0}^{N-1}
\ell_k(x_k,u_k,u_{k+1})
&&
\hspace{11mm}\boxed{\scalebox{1.1}{\textbf{OCP}}}
\label{eq:ocp:cost}
\\
\text{s.t.}
\ 
&x_{k+1}=f_{\bar{\theta}}(x_k,u_k,u_{k+1}), &&k\in \untilwithzero{N{-}1},
\label{eq:ocp:dynamics}
\\
&x_0=x_{\text{init}},
\label{eq:ocp:initial}
\\
&u_{\mathrm{min}}\leq u_k\leq u_{\mathrm{max}},
&&k\in\until{N},
\label{eq:ocp:ubounds}
\\
&e_{\mathrm{min},k}\leq e_k \leq e_{\mathrm{max},k},
&&k\in\until{N}.
\label{eq:ocp:ebounds}
\end{align}
\end{subequations}
The quality of solutions to \ocp depend on the accuracy of the model $f_{\bar\theta}$ parameterized by the last-layer parameters $\bar\theta$:
$$
f_{\bar\theta}(x_k,u_k,u_{k+1})=h(x_k,u_k) + g(z_k; \bar{\theta}),
$$
where $g_i(z_k; \bar{\theta})=\bar\theta_i^\top\phi_i(z_k)$ for each $i=1,\dots,n$. 

Directly deploying an MPC controller that recursively solves \ocp using the prior model parameters $\bar\theta_{i,0}$ after offline meta-training may lead to poor performance in practice, even if one adapts the model online during the task, see Sec.~\ref{sec:results}. 
These observations motivate an active data collection approach to refine the model prior to deployment in challenging applications. Since data collection on a vehicle may be time-consuming and expensive, this raises the question: \textit{What data should we collect to best identify the unknown dynamics?} 

The key insight in answering this question is that learning the model as fast as possible amounts to maximizing the \textit{information gain} from future observations, i.e., to maximizing the \textit{mutual information} between observations and the true system (see \cite[Chapter~2]{cover1999elements} and \cite[Chapter~8]{MacKay2003}). 
The mutual information between two random variables characterizes their mutual dependence and provides a quantitative measure for selecting data that is most informative in identifying the true dynamics~\eqref{eq:dynamics}. Thus, to steer the system towards regions of the state-input space with high model uncertainty, we define the information gathering objective
\begin{equation}\label{eq:info_objective}
\ell_{\text{info}}(x,u):= \frac{1}{2}\sum_{k=0}^{N-1} \sum_{i=1}^n \log(1 + \phi_{i,k}^\top \Lambda_{i,0}^{-1} \phi_{i,k}),
\end{equation}
where $\phi_{i,k}:=\phi_i(x_k, u_k, u_{k+1})$. Using $\ell_{\text{info}}$, we define  the information gathering optimal control problem
\begin{align}
\min_{x,u}
\  \,
&
{-} \alpha \ell_{\text{info}}(x,u) + 
\textstyle\sum_{k=0}^{N-1}\ell_k(x_k,u_k,u_{k+1})
&&
\hspace{0mm}\boxed{\scalebox{1.1}{\textbf{Info-OCP}}} \label{eq:info-ocp}\nonumber
\\
\text{s.t.}
\ \ 
& 
\eqref{eq:ocp:dynamics}-\eqref{eq:ocp:ebounds}, 
\end{align}
where $\alpha > 0$ weighs the information gain compared to the nominal MPC objective. The nominal costs $\ell_k$ encourage smooth control inputs, and the constraints \eqref{eq:ocp:dynamics}-\eqref{eq:ocp:ebounds} ensure that the vehicle remains on the track during data collection.

The information gathering objective \eqref{eq:info_objective} is derived in \cite{lew2022safe} by leveraging the last-layer uncertainty representation of the model, $g$, which yields a closed-form expression of the one-step mutual information between an observation and the dynamics. Similar objectives are used for active learning in \cite{Krause2007,zimmer2018safe,Buisson2020}. As discussed in \cite{lew2022safe}, $\ell_{\mathrm{info}}$ is an approximation of the true total information gain over a trajectory, which is not exactly the sum of the expected information gains per timestep. We found that incorporating the linear regression updates \eqref{eq:update} in the objective (by defining the cost $\tilde{\ell}_{\text{info}}=\frac{1}{2}\sum_{k=0}^{N-1} \sum_{i=1}^n \log(1 + \phi_{i,k}^\top \Lambda_{i,k}^{-1} \phi_{i,k})$ instead) %
does not result in a substantial difference in computed trajectories, while resulting in a more challenging numerical resolution of the resulting information gathering problem.

The information gathering objective \eqref{eq:info_objective} %
guides data collection toward regions of the feature space that enable rapid adaptation. As shown in Fig.~\ref{fig:framework}, the procedure of generating information gathering trajectories and adapting the model on them can be iterated as many times as needed to lower estimates of uncertainty. In our experiments, we found that adaptation on only one or two information gathering trajectories is sufficient for performing reliable drifting maneuvers, as shown in the next section.

\section{Results}\label{sec:results}
We validate the proposed active information gathering framework on a Toyota Supra. The states are estimated at a rate of 100Hz by an OxTS Inertial Navigation System~\cite{oxts2022}, and the MPC problems are solved at approximately $100\mathrm{Hz}$ on an on-board computer equipped with a 3.30GHz Intel Xeon E-2278GE CPU using a standard line search sequential quadratic programming method~\cite[Chapter~18]{nocedal1999numerical} where quadratic programming subproblems are solved with the OSQP solver~\cite{stellato2020osqp}. 
Reference trajectories for drifting are generated by the method in~\cite{Goh2016}. All videos of our experiments are available at \url{https://tinyurl.com/d4pmh8kc}.

\subsection{What do information-rich trajectories look like?}\label{sec:results:info_gathering}

\begin{figure}[t!]
    \centering    \begin{tikzpicture}
    \node[inner sep=0pt] (image) at (0,0)
    {\includegraphics[width=1\linewidth, trim={20cm 14cm 4cm, 14cm}, clip]{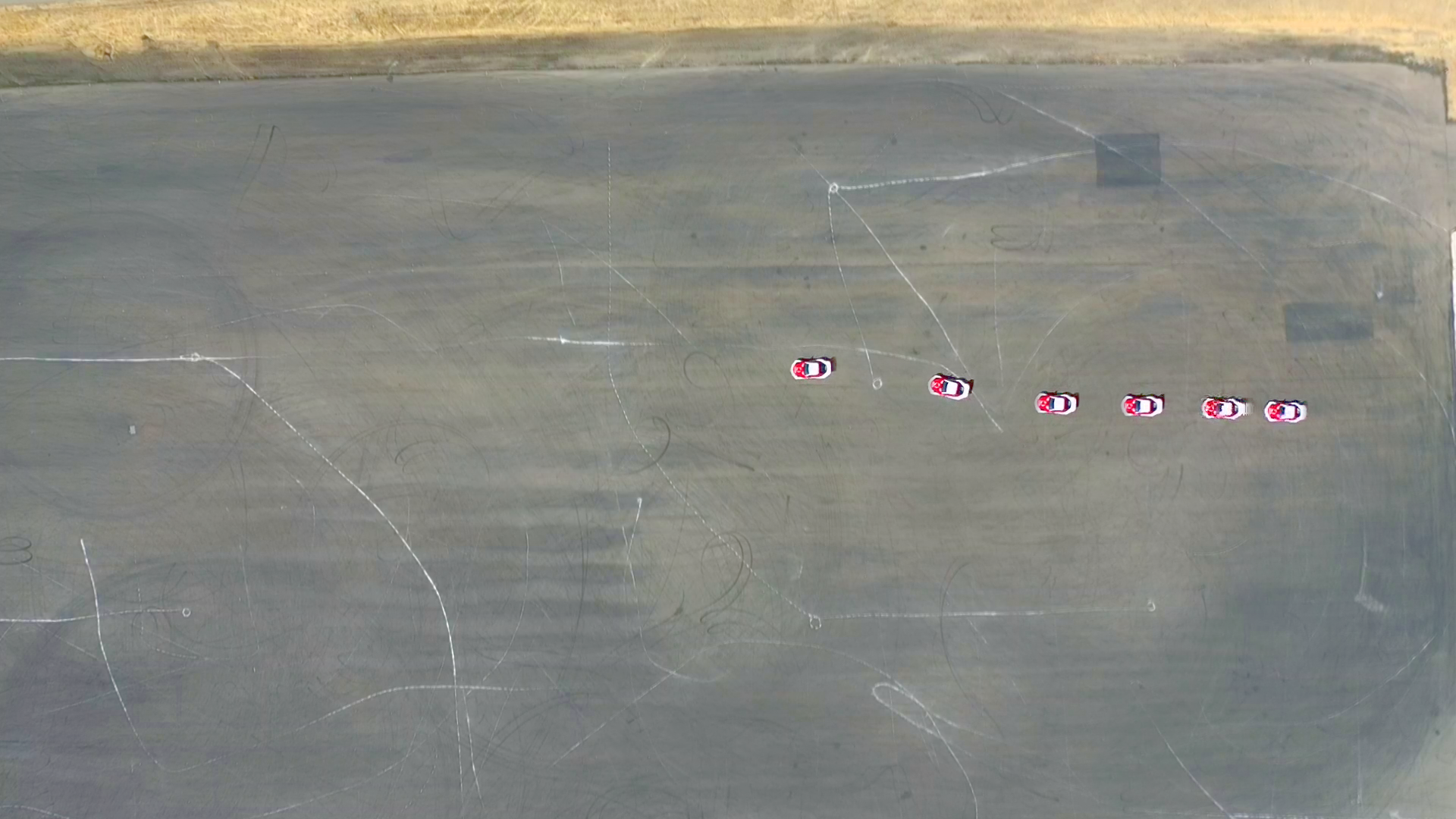}};
    \node[white,anchor=west] at ([xshift=0.55cm, yshift=-0.6cm]image.north west) {
    \textbf{OCP}
    };
    \end{tikzpicture}
    \\[0.5mm]
    \begin{tikzpicture}
        \node[inner sep=0pt] (image) at (0,0)
        {\includegraphics[width=1\linewidth, trim={24cm 14cm 0 14cm}, clip]{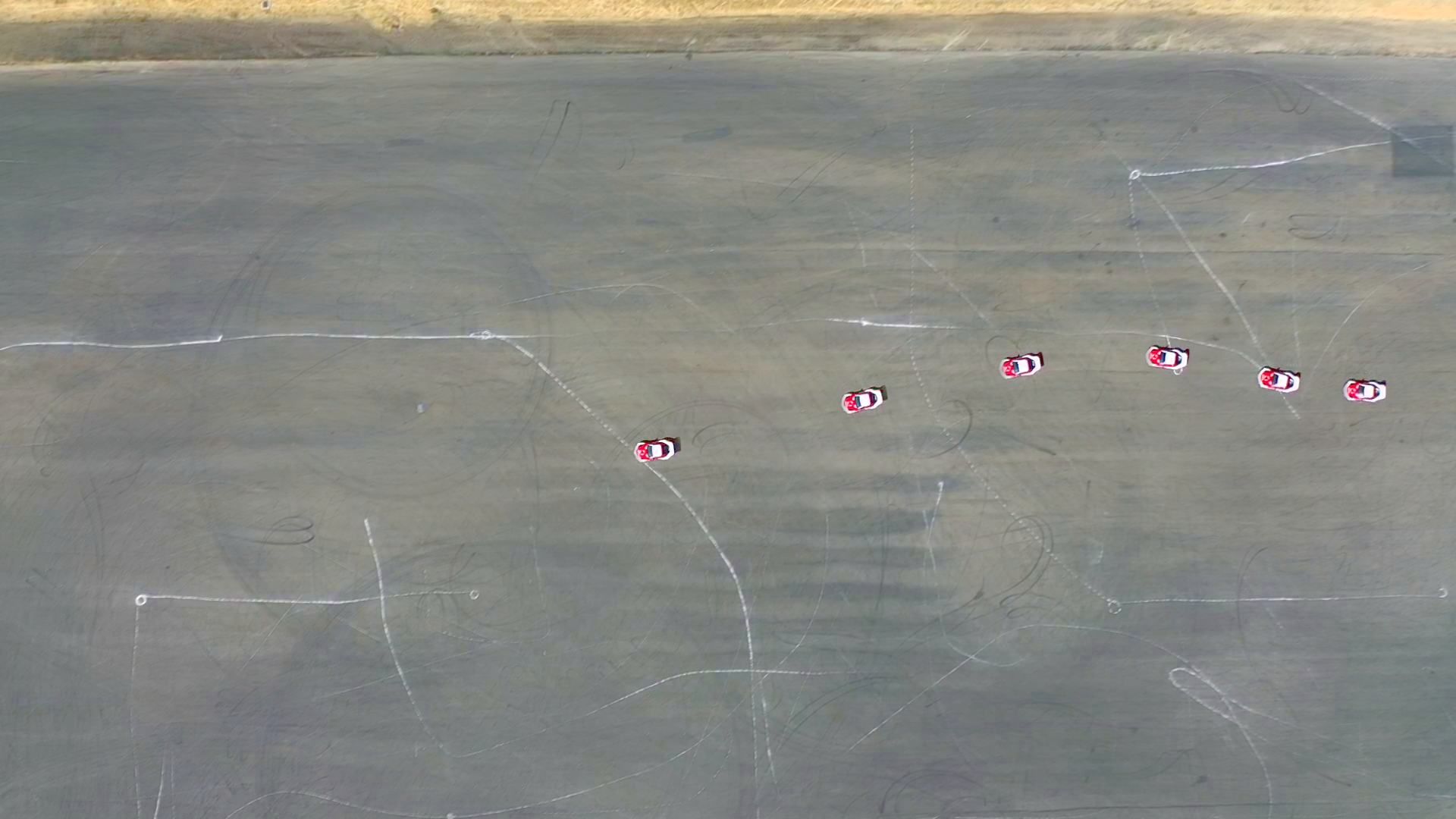}};
        \node[white,anchor=west] at ([xshift=0.55cm, yshift=-0.6cm]image.north west) {
    \textbf{Info-OCP}
    };
    \end{tikzpicture}
    \\[0.5mm]
    \begin{tikzpicture}
        \node[inner sep=0pt] (image) at (0,0)
        {\includegraphics[width=1\linewidth, trim={21cm 18cm 3cm 11cm}, clip]{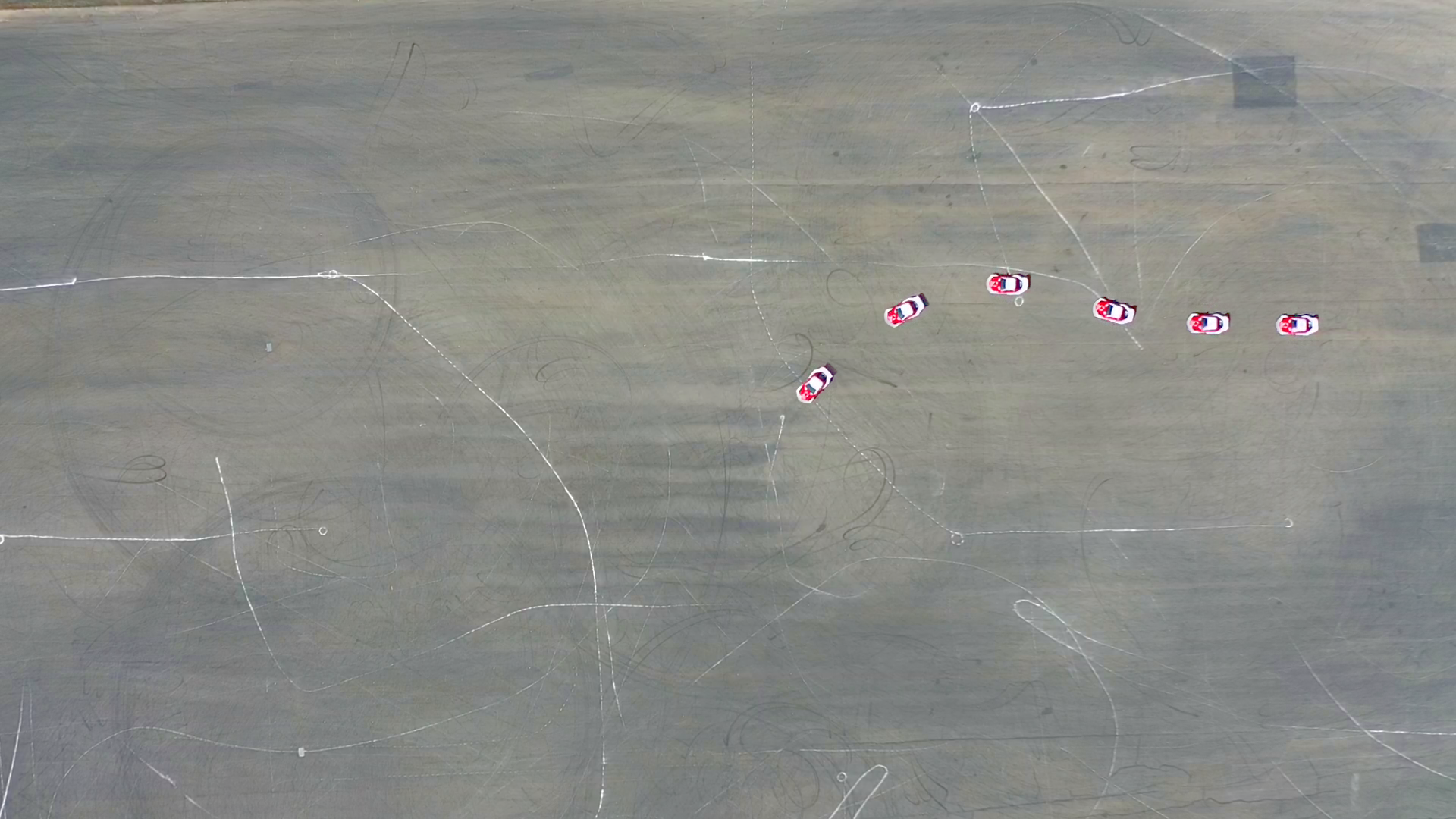}};
        \node[white,anchor=west] at ([xshift=0.55cm, yshift=-0.6cm]image.north west) {
    \textbf{Second Info-OCP}
    };
    \end{tikzpicture}
    \\[2mm]
    \includegraphics[width=1\linewidth, trim={0 0 0 0.1cm}, clip]{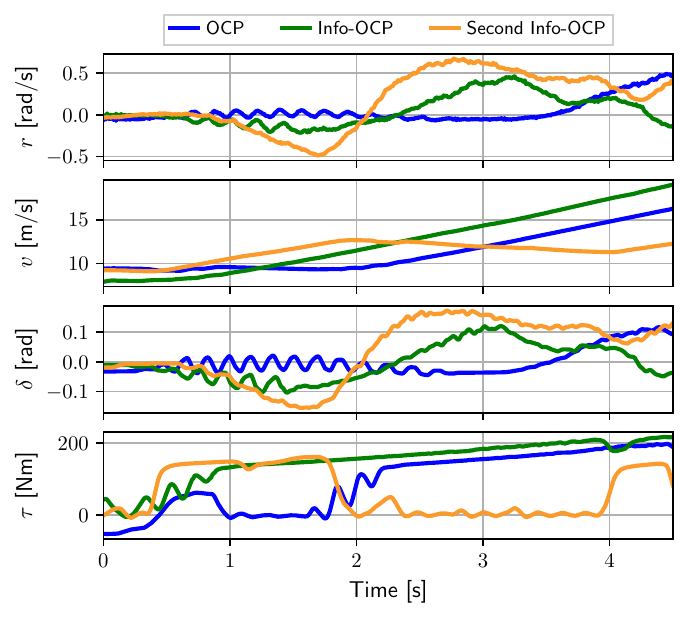}
    \vspace{-16pt}
    \caption{
    Nominal and information gathering trajectories. (Top) Overhead view of the trajectories executed on the vehicle and (Bottom)  corresponding yaw rate, velocity, and control inputs. 
    The duration of all executed trajectories is $4.5$ seconds.
    \vspace{-3mm}}\label{fig:overhead-infos}
\end{figure}
\begin{figure}[t!]
    \centering
    \vspace{-10pt}
    \includegraphics[width=1\linewidth]{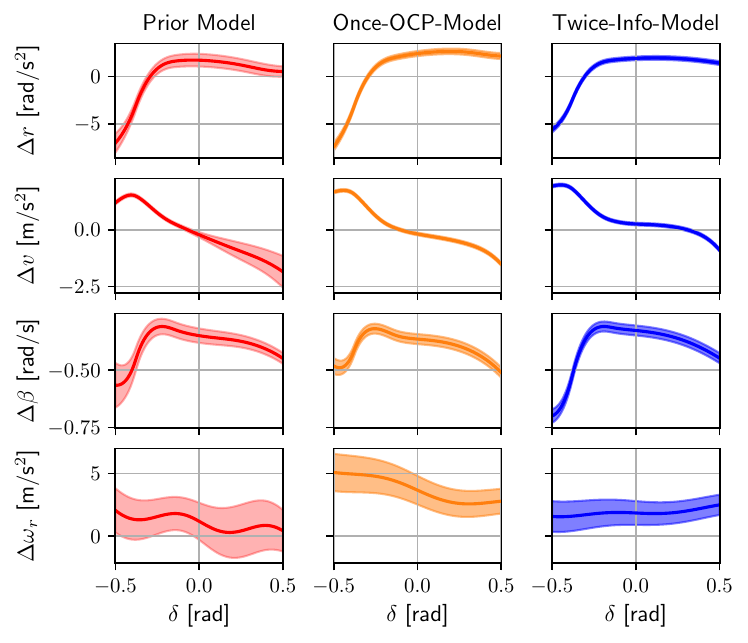}
    \vspace*{-20pt}

    \caption{Predictions of the model pre- and post-adaptation on the information gathering trajectories (see Fig.~\ref{fig:overhead-infos}).
    Mean predictions $\pm 3$ standard deviations for different steering angles $\delta_k = \delta_{k+1}$ at $(r_k,v_k,\beta_k,\omega_{r,k},\tau_k,\tau_{k+1})=(0.9,14.5,-0.45,21,200,200)$. $\Delta r$ denotes $(r_{k+1} - r_k)/\Delta t$ and similarly for other state variables. 
    \vspace{-3mm}
    }
    \label{fig:uncertainties}
\end{figure}

We compare solutions to \ocp with those to \infoocp. We define the problems with $x_{\text{init}}=(0,7.5,0,7.5,0,0)$, a straight-line reference trajectory $x_{\text{ref}}$ from $x_{\text{init}}$ to $x_{\text{ref},N}=(0,20,0,20,0,0)$, and a planning horizon $N=45$, corresponding to $4.5\textrm{s}$-long trajectories. Solutions to \ocp and \infoocp are executed on the vehicle using a low-level tracking controller. We record the trajectories and use them to adapt the Bayesian vehicle model using~\eqref{eq:update}. We refer to (i) the posterior model adapted on the nominal trajectory from \ocp as \modelone and (ii) the posterior model adapted on the information gathering trajectory from \infoocp as \modeltwo. We then use Once-Info-Model to solve \infoocp again using the adapted model parameters to generate a second information gathering trajectory. We refer to the posterior model adapted on both the first and second information gathering trajectories as \modelthree. Vehicle trajectories are shown in Figs.~\ref{fig:overhead-infos}.
Table~\ref{table:covariances} shows the model covariance values $\|\Lambda_i^{-1}\|_2$ prior to adaptation and after adaptation on each trajectory, and Fig.~\ref{fig:uncertainties} shows the models' predictions for a relevant region of the statespace.

\begin{table}[t]
\begin{center}
\caption{Covariance norms $\|\Lambda_i^{-1}\|_2$ for the prior model, %
\modelone (1-OCP-Model), \modeltwo (1-Info-Model), and \modelthree (2-Info-Model).
}
\setlength{\tabcolsep}{5pt}
\begin{tabular}{c|cccc}
\toprule
$\|\Lambda_{i}^{-1}\|_2$ & Prior  & 1-OCP-Model & 1-Info-Model & 2-Info-Model \\ \midrule
$r$      & 3.05  & 1.43   & 0.81    & 0.49    \\ %
$v$      & 27.47 & 1.36   & 1.56    & 0.52    \\ %
$\beta$   & 1.89  & 0.56   & 0.44    & 0.31    \\ %
$\omega_r$     & 1.57  & 0.43   & 0.47    & 0.39    \\ \midrule%
Total        & 33.98 & 3.79   & 3.24    & 1.71    \\ \bottomrule
\end{tabular}
\label{table:covariances}
\end{center}
\end{table}

In Fig.~\ref{fig:overhead-infos}, we see that although the model is meta-trained only on drifting data, solutions to \infoocp are not exactly drifting trajectories. Solutions to \ocp roughly follow the straight-line reference trajectory, whereas the solution to \infoocp applies larger engine torques and steers in each direction to excite the dynamics and provide informative data: Table~\ref{table:covariances} shows that following the solution to \infoocp helps identify dynamics quicker than straight-line driving. Finally, after adapting on the first informative trajectory, the solution to \infoocp applies even larger engine torques and steers the vehicle in each direction to reach larger yaw rates (see Fig.~\ref{fig:overhead-infos}). This helps to reduce uncertainty over the yaw rate and sideslip dynamics (see Table~\ref{table:covariances}). Intuitively, reaching higher yaw rates may help saturate tire forces and identify uncertain tire properties. Although solutions to \infoocp are task-agnostic, identifying tire dynamics may be key to subsequent deployment in dynamic drifting tasks.

\begin{figure}[t!]
\includegraphics[width=\linewidth, trim={0cm 0 0.1cm 0},clip]{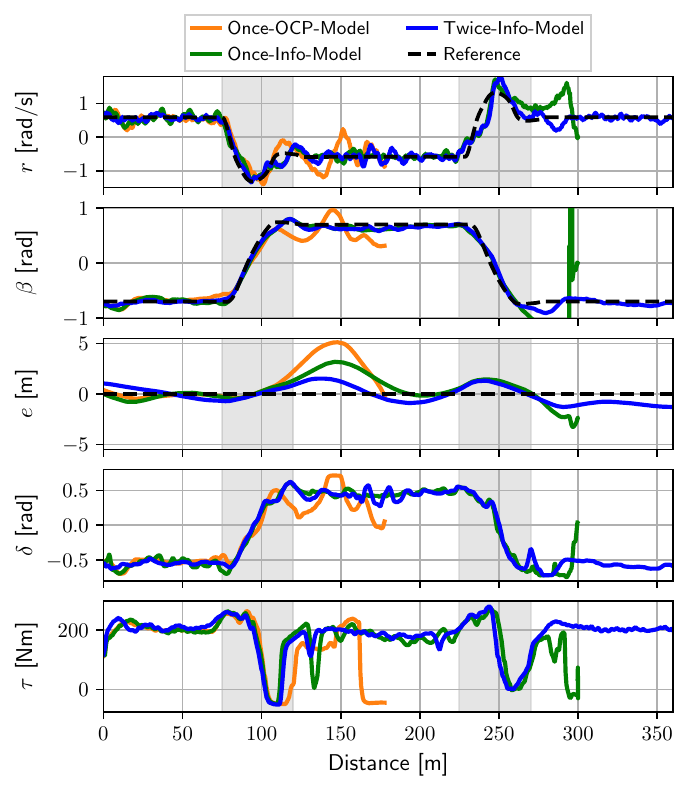}
    \vspace*{-20pt}
    
    \caption{Median Figure-8 drifting performance for 
    the prior model after adapting on the nominal and informative trajectories (Fig.~\ref{fig:overhead-infos}).
    Online adaptation is disabled for all models. 
    Transitions are highlighted in grey. Only \modelthree successfully completes the full maneuver, also shown overhead in Fig.~\ref{fig:main:figure8}.
    \vspace{-2mm}
    }\label{fig:figure8s}
\end{figure}

\subsection{Information gathering enables reliable drifting}\label{sec:fig8}

Next, we consider the task of drifting a Figure-8 trajectory, see Fig.~\ref{fig:main:figure8}. This maneuver is challenging due to the fast and unstable transients %
during transitions between the circular arcs of the trajectory, shown in grey in Fig.~\ref{fig:figure8s}. %

We evaluate the performance of three models for MPC: 
\modelone, \modeltwo, and \modelthree.
To assess the value of the proposed information gathering procedure in isolation, we do not perform online adaptation in this experiment. Each model is given three attempts to drift along the Figure-8, and we present the median tracking performance of each model in Fig.~\ref{fig:figure8s} (additional statistics are reported in Appendix~\ref{app:rmse}). Tracking performance is measured in terms of the root mean squared lateral error along the reference.

We observe that \modelone is unable to complete the drift and deviates too far from the reference trajectory after the first transition. 
This undesirable large deviation is a result of the car transitioning from slipping to gripping after the first transition and then being unable to re-initiate drifting. In contrast, \modeltwo successfully completes the first transition but spins out after the second transition, whereas \modelthree is able to complete the full Figure-8 maneuver, succeeding every time.

Performing at least one round of information gathering 
before deployment helps in completing the first transition. We hypothesize 
that this is because adapting on trajectories from \infoocp results in better estimates of yaw rate dynamics compared to adapting on the nominal \ocp trajectory.
Indeed, Fig.~\ref{fig:overhead-infos} shows that the \infoocp trajectories excite yaw rate dynamics more significantly than the \ocp trajectory does. 
Moreover, Table~\ref{table:covariances} shows that adapting on \infoocp trajectories results in smaller yaw rate covariance norms compared to adapting on the nominal \ocp trajectory, suggesting more accurate yaw rate dynamics estimates after adapting on \infoocp trajectories. 

\subsection{Online adaptation may not be enough for reliable control}\label{sec:online-adaptation-isn't-enough}
\begin{figure}[t!]
    \centering
    \begin{tikzpicture}
    \node[inner sep=0pt] (image) at (0,0)
    {\includegraphics[width=1\linewidth, trim={10cm 0 14cm 0},clip]{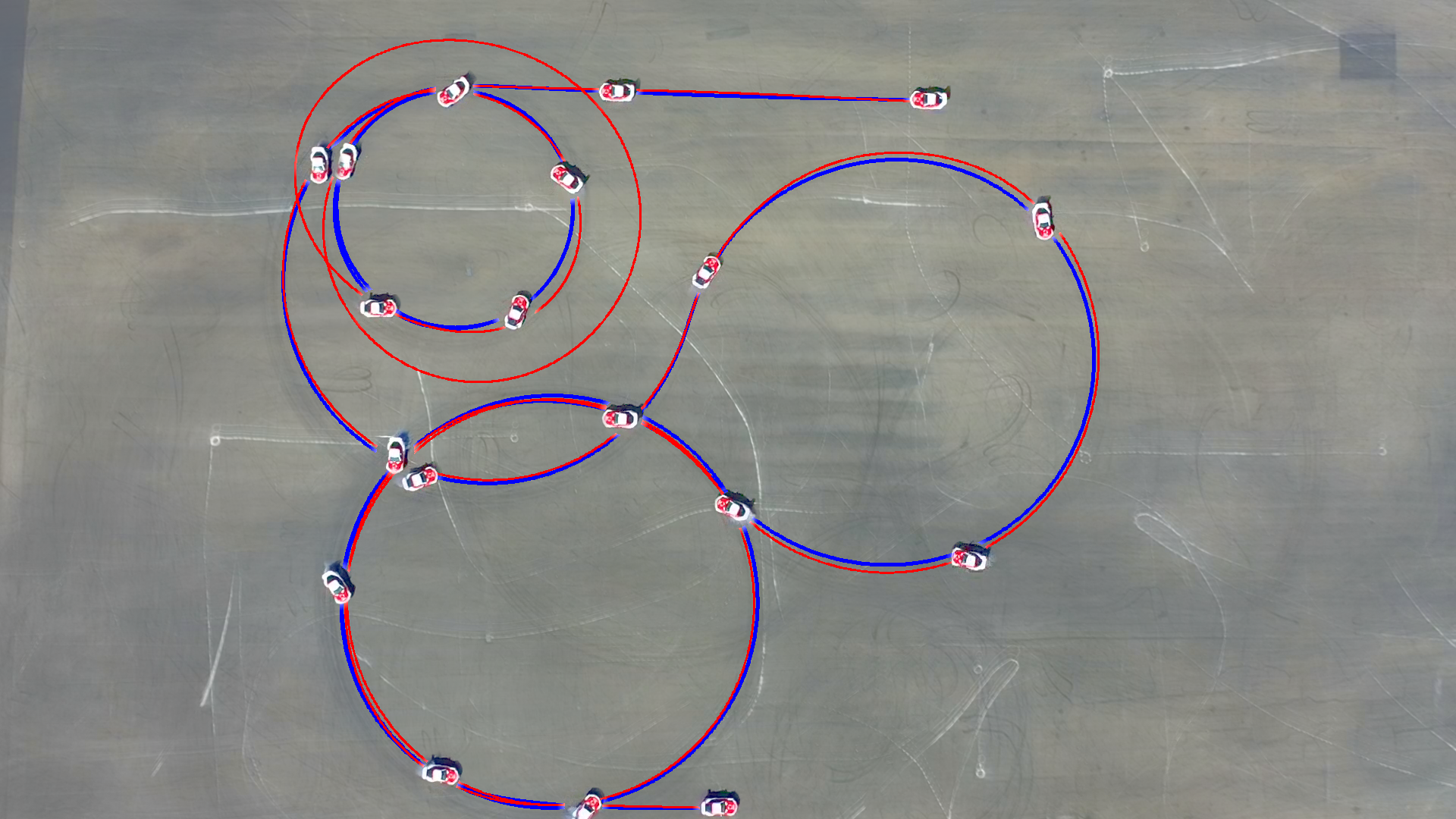}};
    \node[white] at ([xshift=-1.8cm, yshift=0.6cm]image.south east) {
    \begin{tabular}{l}
    \hwplotRed$\;${\footnotesize Online Adaptation} \\
    \hwplotBlue$\;${\footnotesize \text{Twice-Info-Model}} \end{tabular}
    };
\end{tikzpicture}
 \\
\includegraphics[width=1\linewidth, trim={0 0 0 0.0cm},clip]{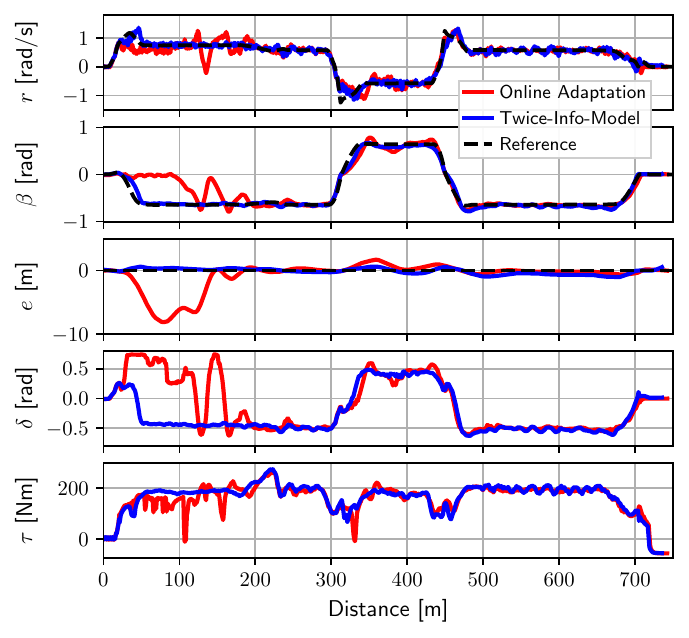}
\vspace{-20pt}
    \caption{
    Best performance in the loops maneuver for the prior model with online adaptation and the model adapted on informative trajectories.
    (Top) Overhead view and (Bottom) time-series data of key signals. Online adaptation is disabled in the Twice-Info-Model.}\label{fig:tandem-overhead}
    \vspace{-10pt}
\end{figure}

Having established the utility of the information gathering and offline adaptation on trajectories generated via \infoocp, we now evaluate the efficacy of online adaptation of the model through the last-layer update rule in~\eqref{eq:update} %
on a loops maneuver
%
shown in Fig.~\ref{fig:tandem-overhead}.  
This maneuver is challenging as the drift is initiated by the MPC controller, in contrast to the hand-brake initiation done in the Figure-8 maneuver. Thus, to successfully initiate the drift, it is essential that the model precisely describes the challenging initial transition between gripping and sliding. %

We compare the performance of 
the prior model adapting online to
the model previously adapted on the two information gathering trajectories from Sec.~\ref{sec:results:info_gathering}. We test each of these models twice and report the best performance of each model in Fig.~\ref{fig:tandem-overhead}. 

Fig.~\ref{fig:tandem-overhead} shows that the model with online adaptation
enables accurate tracking of this challenging trajectory. 
However, while it successfully completes the maneuver and eventually yields accurate tracking, MPC with online adaptation alone leads to large initial errors, requiring upward of $150$ meters to initiate the drift properly. %
This poor transient performance may be problematic for safety-critical applications of controllers for driving at the limits, e.g., for emergency obstacle avoidance \cite{Zhao2022}. 

In contrast, using the model that has adapted offline on the information gathering trajectories enables seamlessly drifting across all transitions, as shown in the in the plots of the tracking errors in position, yaw rate, and sideslip angle in Fig.~\ref{fig:tandem-overhead}. We postulate that the prior model does not capture the transition between gripping and sliding accurately enough to properly initiate the drift. Although the model is meta-trained to be capable of rapid adaptation with little data, results show that online adaptation alone from the prior model does not suffice to initiate the drift seamlessly and leads to a large transient response. %
Information-rich data collection prior to deployment clearly helps in reliably performing this difficult maneuver while minimizing the transient response.

\section{Conclusion}
We present an information gathering framework to actively and efficiently identify nonlinear dynamics for reliable subsequent deployment in MPC. Our approach hinges on an expressive vehicle dynamics model leveraging meta-learned features and Bayesian last-layer weight updates for rapid adaptation. Our results show that collecting informative data first unlocks reliable few-shot deployment within an MPC framework, and we expect that these findings translate to other challenging tasks that involve controlling unstable  %
systems where small model errors may result in large tracking errors. %
This methodology is especially valuable in applications where passive online adaptation %
may not be fast enough to correct the initial model mismatch to complete the task successfully, precisely because unstable systems are difficult to simultaneously learn and control.

In future work, the uncertainty estimates encoded in the Bayesian model could be used in the MPC formulation to robustify constraints during exploration and task deployment \cite{lew2022safe}. These uncertainty estimates could also be used to identify when a sufficient amount of data is collected or to detect out-of-distribution behavior~\cite{sinha2022system}. Finally, although the proposed method is task-agnostic and already demonstrates reliable performance in challenging drifting tasks, leveraging control-oriented identification tools \cite{Richards2023,wagenmaker2024optimal,anderson2024,Lee2024} may 
further enhance the performance of the method and further decrease the data requirements for information gathering.

\appendices

\section{Single-Track Bicycle Dynamics}\label{app:bike}
The nominal dynamics model in \eqref{eq:model}  consists of a single-track bicycle model (see e.g. \cite{goh2024beyond,ding2024drifting}) given by 

{
\small
\begin{equation*}
\begin{bmatrix}
\dot{r}
\\[1mm]
\dot{v}
\\
\\
\dot{\beta}
\\
\\
\dot{\omega}_r
\\[1mm]
\dot{e}
\\
\Delta\dot{\varphi}
\end{bmatrix}
=
\begin{bmatrix}
(
a
	F_{yf}\cos(\delta)
+
	aF_{xf}\sin(\delta)
 - 
b F_{yr})/I_z
\\
\big({-}F_{yf}\sin(\delta-\beta)+F_{xf}\cos(\delta-\beta)
+
\\
\hspace{20mm}
F_{yr}\sin(\beta)
+
F_{xr}\cos(\beta)
\big)/m
\\
-r +
\big(
F_{yf}\cos(\delta-\beta)+F_{xf}\sin(\delta-\beta)+
\\
\hspace{20mm}
F_{yr}\cos(\beta)-F_{xr}\sin(\beta)\big)/(mv)
\\
(G\tau
-F_{xr}R_\textrm{w})/I_\textrm{w}
\\
v\sin(\Delta\varphi)
\\
\dot\beta+r-\kappa_{\textrm{ref}}v\cos(\Delta\varphi)/(1-\kappa_{\textrm{ref}}e)
\end{bmatrix},
\end{equation*}
}%
where $(a,b)$ are the distances from the center of gravity to the front and rear axles, respectively, $(I_z,m)$ are the vehicle inertia and mass, respectively, $(\subscr{R}{w},\subscr{I}{w})$ are the wheel radius and the rear axle inertia, respectively, $G$ is the gear ratio from engine to rear axle, $\subscr{\kappa}{ref}$ is the curvature of the reference trajectory, and $(F_{yf}, F_{xf}, F_{yr}, F_{xr})$ denote tire forces and are modeled using an isotropic coupled slip brush Fiala model~\cite{Svendenius2007}. Following~\cite{goh2024beyond}, since we test on a rear-wheel drive vehicle, we set $F_{xf} = 0$.

\section{Additional Statistics on Figure-8 Trajectory}\label{app:rmse}
We report additional statistics to compare the performance of \modelone, \modeltwo, and \modelthree for the task of drifting the Figure-8 trajectory from Sec.~\ref{sec:fig8}. In Table~\ref{table:lateral-error-rmse}, we present the root mean square (RMS) lateral deviation $e$ over three trials for each of the three models. We also report the RMS sideslip angle error, $e_{\beta} = \beta - \subscr{\beta}{ref},$ in Table~\ref{table:rmse-sideslip}.
Since neither \modelone nor \modeltwo successfully completed the entirety of the Figure-8, we truncated the RMS computation before the spin out or the manual takeover for these models.
\modelthree always achieves the lowest RMS errors and is also the only model that completes the entirety of the Figure-8 trajectory without ever spinning out. 

\begin{table}[t]
\begin{center}
\caption{Root mean squared (RMS) lateral deviation for \modelone, \modeltwo, and \modelthree over three attempts of drifting the Figure-8 trajectory. Units are [m].  %
}
\setlength{\tabcolsep}{5pt}
\begin{tabular}{c|ccc}
\toprule
$\mathrm{RMS}(e)$ & \modelone & \modeltwo & \modelthree \\ \midrule
Best      & 2.31  & 1.18   & \textbf{0.61}    \\ %
Median      & 3.19 & 1.31   & \textbf{0.83}  \\ %
Worst   & 7.35  & 3.64   & \textbf{1.29}   \\
\bottomrule
\end{tabular}
\label{table:lateral-error-rmse}
\end{center}
\end{table}

\begin{table}[t]
\begin{center}
\caption{Root mean squared (RMS) sideslip angle error $e_\beta = \beta - \beta_{\mathrm{ref}}$ for \modelone, \modeltwo, and \modelthree over three attempts of drifting the Figure-8 trajectory. Units are [rad].  %
}
\setlength{\tabcolsep}{5pt}
\begin{tabular}{c|ccc}
\toprule
$\mathrm{RMS}(e_\beta)$ & \modelone & \modeltwo & \modelthree \\ \midrule
Best      & 0.137  & 0.105   & \textbf{0.068}    \\ %
Median      & 0.261 & 0.348   & \textbf{0.070}  \\ %
Worst   & 0.384  & 0.526   & \textbf{0.084}   \\
\bottomrule
\end{tabular}
\label{table:rmse-sideslip}
\end{center}
\end{table}

\section*{Acknowledgments}
\noindent 
We thank Phung Nguyen, Steven Goldine, and William Kettle for their support with the test platform and experiments.

\bibliographystyle{ieeetr}
\balance
\bibliography{sasha}

\begin{thebibliography}{10}

\bibitem{Stein2003}
G.~Stein, ``Respect the unstable,'' {\em {IEEE Control Systems Magazine}}, vol.~23, no.~4, pp.~12--25, 2003.

\bibitem{velenis2011steady}
E.~Velenis, D.~Katzourakis, E.~Frazzoli, P.~Tsiotras, and R.~Happee, ``Steady-state drifting stabilization of {RWD} vehicles,'' {\em Control Engineering Practice}, vol.~19, no.~11, pp.~1363--1376, 2011.

\bibitem{cutler2016autonomous}
M.~Cutler and J.~P. How, ``Autonomous drifting using simulation-aided reinforcement learning,'' in {\em {Proc.\ IEEE Conf.\ on Robotics and Automation}}, pp.~5442--5448, 2016.

\bibitem{djeumou2023autonomous}
F.~Djeumou, J.~Y.~M. Goh, U.~Topcu, and A.~Balachandran, ``Autonomous drifting with 3 minutes of data via learned tire models,'' in {\em {Proc.\ IEEE Conf.\ on Robotics and Automation}}, pp.~968--974, 2023.

\bibitem{goh2024beyond}
J.~Y.~M. Goh, M.~Thompson, J.~Dallas, and A.~Balachandran, ``Beyond the stable handling limits: {N}onlinear model predictive control for highly transient autonomous drifting,'' {\em Vehicle System Dynamics}, pp.~1--24, 2024.

\bibitem{Weber24}
T.~P. Weber and J.~C. Gerdes, ``Modeling and control for dynamic drifting trajectories,'' {\em IEEE Transactions on Intelligent Vehicles}, vol.~9, no.~2, pp.~3731--3741, 2024.

\bibitem{Hewing2020}
L.~Hewing, K.~P. Wabersich, M.~Menner, and M.~N. Zeilinger, ``Learning-based model predictive control: Toward safe learning in control,'' {\em Annual Review of Control, Robotics, and Autonomous Systems}, vol.~3, no.~1, pp.~269--296, 2020.

\bibitem{kabzan2019learning}
J.~Kabzan, L.~Hewing, A.~Liniger, and M.~N. Zeilinger, ``Learning-based model predictive control for autonomous racing,'' {\em IEEE Robotics and Automation Letters}, vol.~4, no.~4, pp.~3363--3370, 2019.

\bibitem{spielberg2021neural}
N.~A. Spielberg, M.~Brown, and J.~C. Gerdes, ``Neural network model predictive motion control applied to automated driving with unknown friction,'' {\em IEEE Transactions on Control Systems Technology}, vol.~30, no.~5, pp.~1934--1945, 2021.

\bibitem{djeumou2024one}
F.~Djeumou, T.~Lew, N.~Ding, M.~Thompson, M.~Suminaka, M.~Greiff, and J.~Subosits, ``One model to drift them all: {P}hysics-informed conditional diffusion model for driving at the limits,'' in {\em {Conf.\ on Robot Learning}}, 2024.

\bibitem{ding2024drifting}
N.~Ding, M.~Thompson, J.~Dallas, J.~Y.~M. Goh, and J.~Subosits, ``Drifting with unknown tires: {L}earning vehicle models online with neural networks and model predictive control,'' in {\em IEEE Intelligent Vehicles Symposium}, pp.~2545--2552, 2024.

\bibitem{Brunke2022}
L.~Brunke, M.~Greeff, A.~W. Hall, Z.~Yuan, S.~Zhou, J.~Panerati, and A.~P. Schoellig, ``Safe learning in robotics: From learning-based control to safe reinforcement learning,'' {\em Annual Review of Control, Robotics, and Autonomous Systems}, vol.~5, no.~1, pp.~411--444, 2022.

\bibitem{Abbasi2011}
Y.~Abbasi-Yadkori, D.~Pal, and C.~Szepesvári, ``Improved algorithms for linear stochastic bandits,'' in {\em {Conf.\ on Neural Information Processing Systems}}, 2011.

\bibitem{wagenmaker2024optimal}
A.~Wagenmaker, G.~Shi, and K.~G. Jamieson, ``Optimal exploration for model-based {RL} in nonlinear systems,'' {\em {Conf.\ on Neural Information Processing Systems}}, 2024.

\bibitem{Koller2018}
T.~Koller, M.~T. F.~Berkenkamp, and A.~Krause, ``Learning-based model predictive control for safe exploration,'' in {\em {Proc.\ IEEE Conf.\ on Decision and Control}}, pp.~6059--6066, 2018.

\bibitem{Buisson2020}
M.~Buisson-Fenet, F.~Solowjow, and S.~Trimpe, ``Actively learning {G}aussian process dynamics,'' in {\em Conf. on Learning for Dynamics and Control}, pp.~5--15, 2020.

\bibitem{mania2022active}
H.~Mania, M.~I. Jordan, and B.~Recht, ``Active learning for nonlinear system identification with guarantees,'' {\em Journal of Machine Learning Research}, vol.~23, no.~32, pp.~1--30, 2022.

\bibitem{saviolo2023active}
A.~Saviolo, J.~Frey, A.~Rathod, M.~Diehl, and G.~Loianno, ``Active learning of discrete-time dynamics for uncertainty-aware model predictive control,'' {\em IEEE Transactions on Robotics}, vol.~40, pp.~1273--1291, 2024.

\bibitem{Zhang2021}
W.~Zhang, M.~Tognon, L.~Ott, R.~Siegwart, and J.~Nieto, ``Active model learning using informative trajectories for improved closed-loop control on real robots,'' in {\em {Proc.\ IEEE Conf.\ on Robotics and Automation}}, pp.~4467--4473, 2021.

\bibitem{lew2022safe}
T.~Lew, A.~Sharma, J.~Harrison, A.~Bylard, and M.~Pavone, ``Safe active dynamics learning and control: {A} sequential exploration--exploitation framework,'' {\em {IEEE Transactions on Robotics}}, vol.~38, no.~5, pp.~2888--2907, 2022.

\bibitem{Dean2019}
S.~Dean, H.~Mania, N.~Matni, B.~Recht, and S.~Tu, ``On the sample complexity of the linear quadratic regulator,'' {\em Foundations of Computational Mathematics}, vol.~20, no.~4, pp.~633--679, 2019.

\bibitem{Mania2019}
H.~Mania, S.~Tu, and B.~Recht, ``Certainty equivalence is efficient for linear quadratic control,'' in {\em {Conf.\ on Neural Information Processing Systems}}, 2019.

\bibitem{kim2023bridging}
T.~Kim, J.~Mun, J.~Seo, B.~Kim, and S.~Hong, ``Bridging active exploration and uncertainty-aware deployment using probabilistic ensemble neural network dynamics,'' in {\em {Robotics: Science and Systems}}, 2023.

\bibitem{sukhija2024optimistic}
B.~Sukhija, L.~Treven, C.~Sancaktar, S.~Blaes, S.~Coros, and A.~Krause, ``Optimistic active exploration of dynamical systems,'' {\em Advances in Neural Information Processing Systems}, pp.~38122--38153, 2024.

\bibitem{harrison2020meta}
J.~Harrison, A.~Sharma, and M.~Pavone, ``Meta-learning priors for efficient online {B}ayesian regression,'' in {\em Algorithmic Foundations of Robotics XIII: Proceedings of the 13th Workshop on the Algorithmic Foundations of Robotics}, pp.~318--337, 2020.

\bibitem{lee2019meta}
K.~Lee, S.~Maji, A.~Ravichandran, and S.~Soatto, ``Meta-learning with differentiable convex optimization,'' in {\em {IEEE/CVF Conf.\ on Computer Vision and Pattern Recognition}}, pp.~10657--10665, 2019.

\bibitem{bertinetto2019meta}
L.~Bertinetto, J.~F. Henriques, P.~H.~S. Torr, and A.~A.~Vedaldi, ``Meta-learning with differentiable closed-form solvers,'' in {\em International Conference on Learning Representations}, 2019.

\bibitem{Richards2023}
S.~M. Richards, N.~Azizan, J.-J. Slotine, and M.~Pavone, ``Control-oriented meta-learning,'' {\em {Int.\ Journal of Robotics Research}}, vol.~42, no.~10, pp.~777--797, 2023.

\bibitem{o2022neural}
M.~O’Connell, G.~Shi, X.~Shi, K.~Azizzadenesheli, A.~Anandkumar, Y.~Yue, and S.-J. Chung, ``Neural-fly enables rapid learning for agile flight in strong winds,'' {\em Science Robotics}, vol.~7, no.~66, pp.~1--15, 2022.

\bibitem{Banerjee2020}
S.~Banerjee, J.~Harrison, P.~M. Furlong, and M.~Pavone, ``Adaptive meta-learning for identification of rover-terrain dynamics,'' in {\em {Int.\ Symp.\ on Artificial Intelligence, Robotics and Automation in Space}}, 2020.

\bibitem{cover1999elements}
T.~M. Cover, {\em Elements of Information Theory}.
\newblock John Wiley \& Sons, 1999.

\bibitem{MacKay2003}
D.~J.~C. MacKay, {\em Information Theory, Inference, and Learning Algorithms}.
\newblock Cambridge University Press, 2003.

\bibitem{Krause2007}
A.~Krause and C.~Guestrin, ``Nonmyopic active learning of {G}aussian processes,'' in {\em {Int.\ Conf.\ on Machine Learning}}, 2007.

\bibitem{zimmer2018safe}
C.~Zimmer, M.~Meister, and D.~Nguyen-Tuong, ``Safe active learning for time-series modeling with {G}aussian processes,'' in {\em {Conf.\ on Neural Information Processing Systems}}, 2018.

\bibitem{oxts2022}
OxTS, ``User manual: {NCOM} data format for the efficient communication of navigation measurements,'' 2022.

\bibitem{nocedal1999numerical}
J.~Nocedal and S.~J. Wright, {\em Numerical Optimization}.
\newblock Springer, 1999.

\bibitem{stellato2020osqp}
B.~Stellato, G.~Banjac, P.~Goulart, A.~Bemporad, and S.~Boyd, ``{OSQP}: {A}n operator splitting solver for quadratic programs,'' {\em Mathematical Programming Computation}, vol.~12, no.~4, pp.~637--672, 2020.

\bibitem{Goh2016}
J.~Y. Goh and J.~C. Gerdes, ``Simultaneous stabilization and tracking of basic automobile drifting trajectories,'' in {\em IEEE Intelligent Vehicles Symposium}, pp.~597--602, 2016.

\bibitem{Zhao2022}
T.~Zhao, E.~Yurtsever, and G.~Rizzoni, ``Justifying emergency drift control for automated vehicles,'' in {\em IFAC Symposium on Advances in Automotive Control}, vol.~55, pp.~141--148, 2022.

\bibitem{sinha2022system}
R.~Sinha, A.~Sharma, S.~Banerjee, T.~Lew, R.~Luo, S.~M. Richards, Y.~Sun, E.~Schmerling, and M.~Pavone, ``A system-level view on out-of-distribution data in robotics,'' {\em arXiv preprint arXiv:2212.14020}, 2022.

\bibitem{anderson2024}
S.~Anderson and J.~P. Hespanha, ``Control-oriented identification via stochastic optimization,'' {\em IEEE Control Systems Letters}, vol.~8, pp.~1865--1870, 2024.

\bibitem{Lee2024}
B.~D. Lee, I.~Ziemann, G.~J. Pappas, and N.~Matni, ``Active learning for control-oriented identification of nonlinear systems,'' {\em arXiv preprint arXiv:2404.09030}, 2024.

\bibitem{Svendenius2007}
J.~Svendenius, {\em Tire modeling and friction estimation}.
\newblock PhD thesis, Lund University, 2007.

\end{thebibliography}

\end{document}